\documentclass[10pt,twocolumn,letterpaper]{article}

\usepackage{cvpr}
\usepackage{times}
\usepackage{epsfig}
\usepackage{graphicx}
\usepackage{amsmath}
\usepackage{amssymb}
\usepackage{makecell}
\usepackage{algorithm}
\usepackage{algorithmic}
\usepackage{subfigure}
\usepackage{multirow}

\usepackage[pagebackref=true,breaklinks=true,colorlinks,bookmarks=false]{hyperref}

\cvprfinalcopy 

\def\cvprPaperID{2994} 

\ifcvprfinal\fi
\begin{document}

\title{Contrastive Prototype Learning with Augmented Embeddings \\ for Few-Shot Learning}
\author{Yizhao Gao$^1$, Nanyi Fei$^1$, Guangzhen Liu$^1$, Zhiwu Lu$^1$, Tao Xiang$^2$, Songfang Huang$^3$\\
$^1$Renmin University of China, $^2$University of Surrey, $^3$Alibaba Group}


\maketitle
\begin{abstract}
  Most recent few-shot learning (FSL) methods are based on meta-learning with episodic training. In each meta-training episode, a discriminative feature embedding and/or classifier are first constructed from a support set in an inner loop, and then evaluated in an outer loop using a query set for model updating. This query set sample centered learning objective is however intrinsically limited in addressing the lack of training data problem in the support set. In this paper, a novel contrastive prototype learning with augmented embeddings (CPLAE) model is proposed to overcome this limitation. First, data augmentations are introduced to both the support and query sets with each sample now being represented as an augmented embedding (AE) composed of concatenated embeddings of both the original and augmented versions. Second, a novel support set class prototype centered contrastive loss is proposed for contrastive prototype learning (CPL). With a class prototype as an anchor, CPL aims to pull the query samples of the same class closer and those of different classes further away. This support set sample centered loss is highly complementary to the existing query centered loss, fully exploiting the limited training data in each episode. Extensive experiments on several benchmarks demonstrate that our proposed CPLAE achieves new state-of-the-art. 
\end{abstract}


\section{Introduction}

Deep convolutional neural networks (CNNs) \cite{krizhevsky2012nips, he2016resnet} have witnessed tremendous successes in many visual recognition tasks. However, the powerful learning ability of CNNs depends on a large amount of manually labeled training data. In practice, sufficient manual annotation is often too costly and may even be infeasible (e.g., for rare object classes). This has severely limited the usefulness of CNNs for real-world applications. Many attempts have been made recently to mitigate such a limitation from the transfer learning perspective, resulting in the popular research line of few-shot learning (FSL) \cite{Fei-Fei2003iccv, feifei2006pami}. FSL aims to transfer knowledge learned from abundant seen class samples to a set of unseen classes (only with few shots per class).

Most recent FSL methods are based on meta-learning \cite{oriol2016nips, snell2017nips, finn2017icml, sung2018cvpr}. That is, they learn an algorithm or model across a set of sampled FSL training/seen tasks, with the objective of making it generalizable to any unseen test tasks. To that end, an episodic training strategy is adopted, i.e., the seen tasks are arranged into learning episodes, each of which contains $n$ classes and $k$ labeled samples per class to simulate the setting for the unseen test tasks. In each episode, the meta-training data is further split into a support set and a query set. Part of the CNN model (e.g., feature embedding subnet, classification layers, or parameter initialization) to be meta-learned is first obtained in an inner loop using the support set. It is then evaluated in an outer loop using the query set for model updating.  

These meta-learning based FSL methods differ mainly in which part of the model is meta-learned. Among them, those meta-learning a feature embedding or distance metric have dominated the state-of-the-art. Many of them \cite{allen2019imp,li2019dn4,afrasiyabi2020align,ye2020feat,zhang2020deepemd} are based on the prototypical network (ProtoNet) \cite{snell2017nips} for its simplicity and competitive performance with various extensions. Given a feature embedding network learned from the preceding episode, these methods first compute one prototype per class as the support set class mean; these prototypes are then used as a nearest neighbour classifier in the outer loop on the query set to update the feature embedding. In other words, the meta-learning loss is query centred aiming to make sure that each query sample is close to its corresponding class prototype whilst being further away from other prototypes. 

However, this design has severe limitations in addressing a fundamental challenge in FSL, i.e., the lack of support set samples. By definition, each class is only represented by few shots, i.e., $n$ is very small. This problem is actually exacerbated by taking a query-centred only meta-learning loss (with prototype based class representation) that considers the relationship between each query against the $k$ prototypes individually rather than collectively as a distribution.  

\begin{figure}[t]
\centering
\subfigure[ProtoNet]{
\includegraphics[width=0.32\columnwidth]{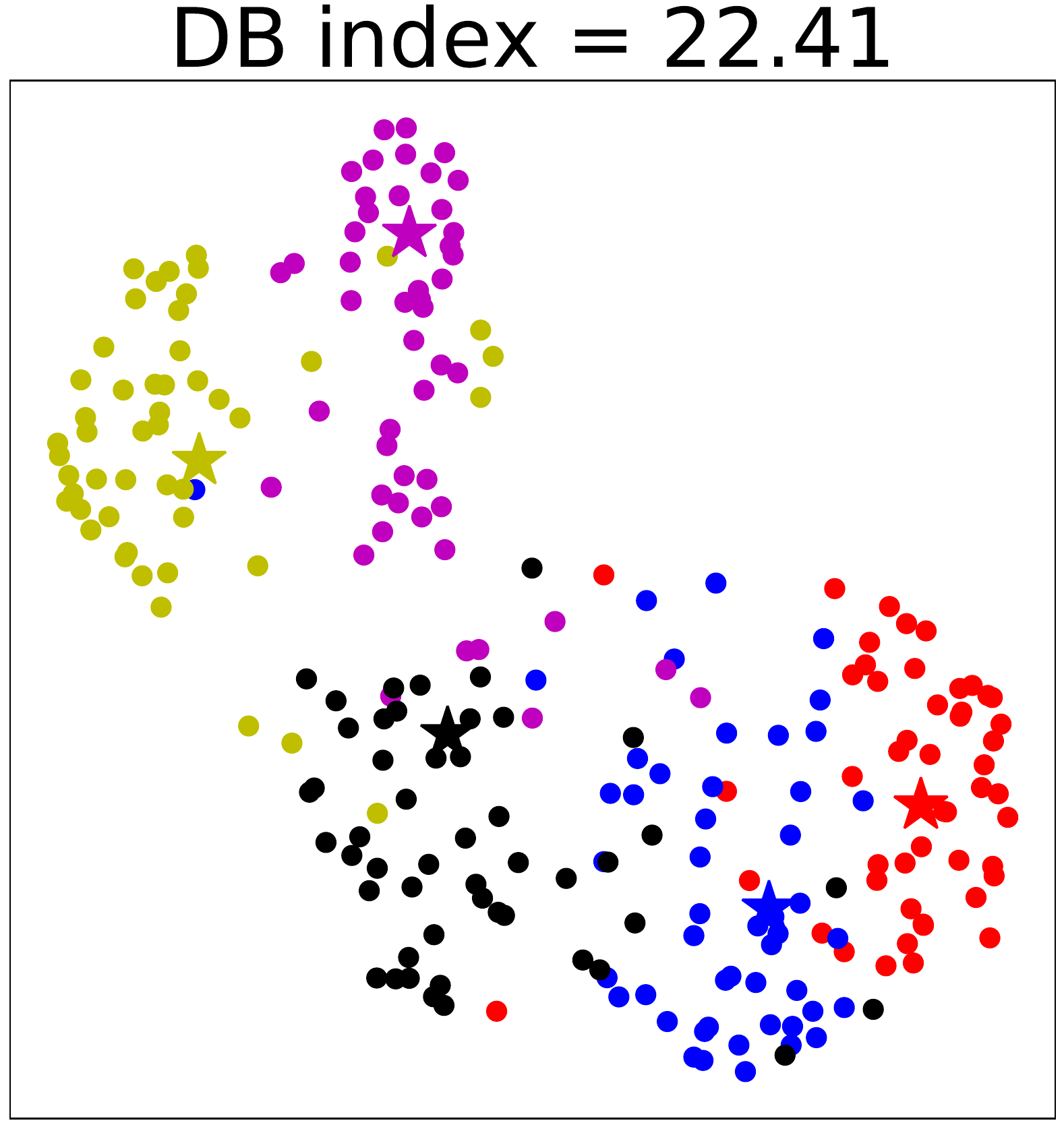}
\label{subfig:proto}
}%
\subfigure[ProtoNet+AE]{
\includegraphics[width=0.32\columnwidth]{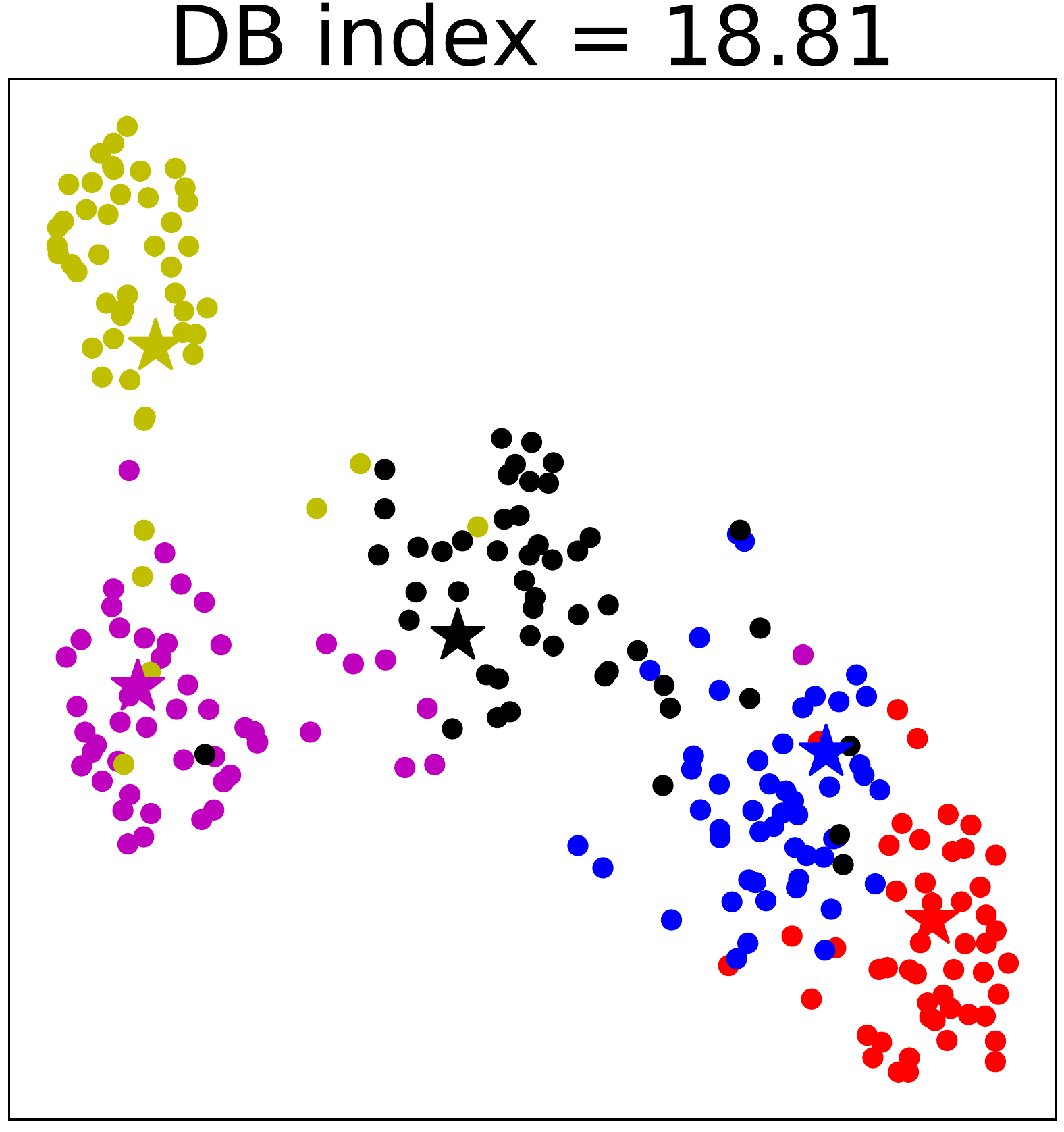}
\label{subfig:proto+AE}
}%
\subfigure[ProtoNet+AE+CPL]{
\includegraphics[width=0.32\columnwidth]{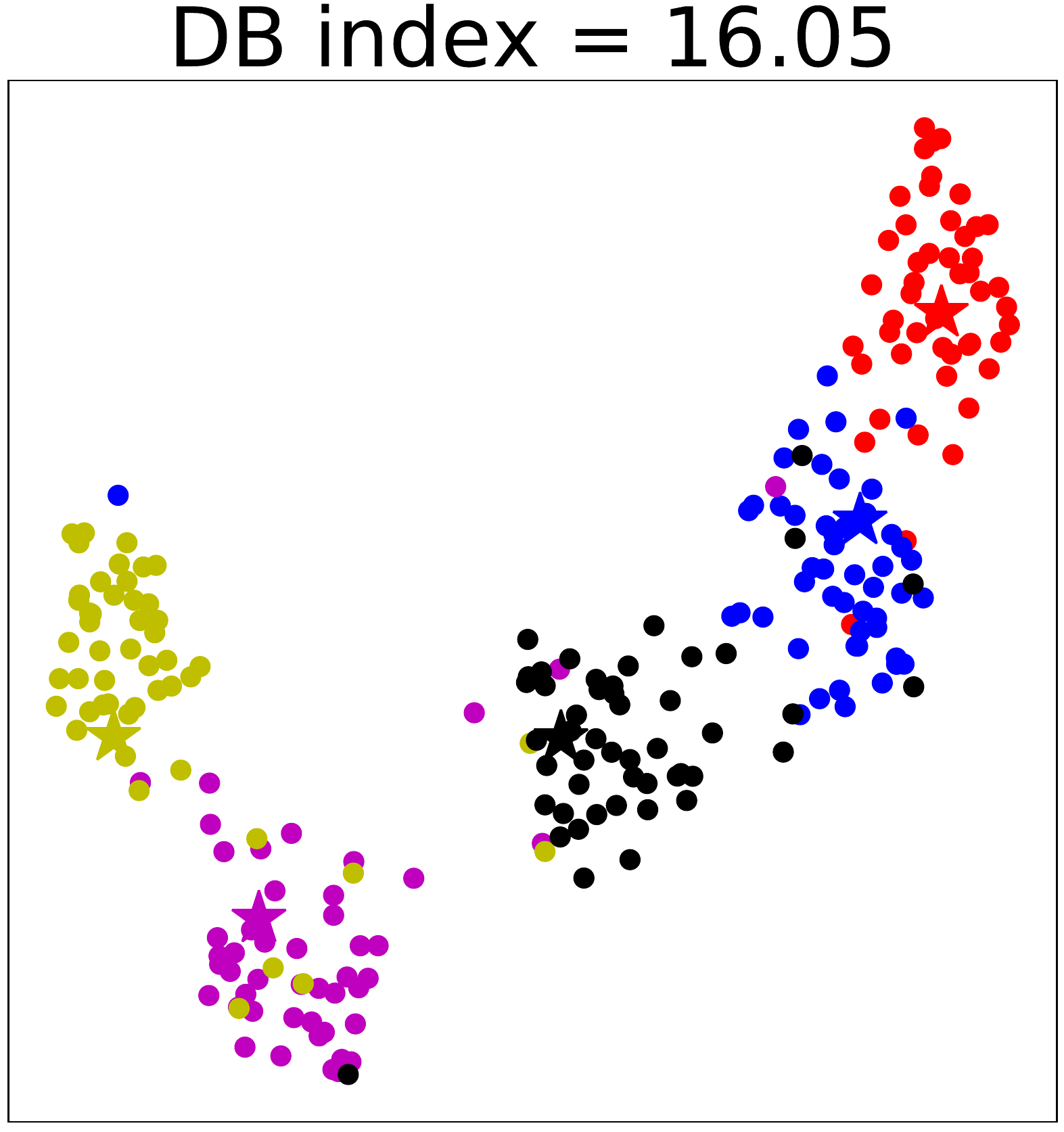}
\label{subfig:proto+AE+CPL}
}
\caption{Feature visualization of the same meta-test episode for three FSL models using the UMAP algorithm \cite{UMAP}. Both ProtoNet and ProtoNet+AE take a query-centered view, while our CPL takes a prototype-centered view. The Davies-Bouldin index (DB index)~\cite{dbindex} is used to measure the intra-class variation, which takes a lower value when the data clustering structure is clearer/better.}
\label{fig:intro}
\vspace{-0.1in}
\end{figure}

In this paper, to address the lack of support set sample problem, we propose a novel contrastive prototype learning with augmented embeddings (CPLAE) model for FSL. Our proposed CPLAE has two new components: \textbf{(1) Augmented embedding (AE)} -- each sample in the support/query set and its three augmented versions are integrated to obtain an augmented embedding. Data augmentation is commonly used in training a CNN for improving its generalization to unseen test data. It has also been considered for FSL \cite{gidaris2019selfsup, su2020when, mangla2020wacv}. Rather than using augmentation for auxiliary tasks as in existing works, we concatenate the  feature embeddings of both the original and augmented versions of each sample to form a richer AE space for meta-learning. \textbf{(2) Contrastive prototype learning (CPL) } -- Similar to \cite{allen2019imp, li2019dn4, afrasiyabi2020align, ye2020feat, zhang2020deepemd}, our CPLAE is also based on ProtoNet for meta-learning a feature embedding. Differently, instead of using only query sample centered learning objectives, we additionally introduce a novel support sample centered loss to make full use of the limited training data in each episode. Our CPL loss is a supervised contrastive loss \cite{Khosla2020SCL} adapted to FSL. More specifically, each prototype  is used as an anchor with query set samples of the same class used as positives and all other query samples as negatives. Contrary and yet complementary to the existing query centered loss that constrains the support set distribution, this support set prototype centered loss regularizes the query set distribution. Combining both losses results in a better embedding space where different classes are more separable (see Figure~\ref{fig:intro}). CPL and AE are integrated seamlessly in our CPLAE in that different AE concatenation orders are applied to the anchor and negatives/positives to further boost the generalization ability of the learned embedding. 

Our main contributions are: (1) For the first time, we identify the limitations of existing embedding-based meta-learning methods in dealing with scarce training samples for FSL, caused by adopting only query centered learning objectives. (2) As a remedy, we propose a novel CPLAE model composed of two components (i.e., AE and CPL). Combining AE and supervised contrastive learning seamlessly, our CPL loss enforces a support set centered constraint on the query set sample distribution, thus being complementary to existing query centered losses and effectively making full use of the limited training data. (3) Extensive experiments on several benchmarks demonstrate that our proposed CPLAE achieves new state-of-the-art.  

\section{Related Work}

\noindent\textbf{Few-Shot Learning.}~Most recent FSL methods follow the meta-learning paradigm. They can be  roughly divided into four groups: (1) Embedding/Metric-based methods learn shared task-agnostic embedding spaces/distance metrics or learn task-specific metrics. The former methods either learn an embedding space where a fixed metric (e.g., cosine \cite{oriol2016nips} or Euclidean distance \cite{snell2017nips}) can be used,  or learn a distance metric (e.g., CNN-based relation modules \cite{sung2018cvpr, wu2019parn}, ridge regression \cite{bertinetto2019iclr}, and graph neural networks \cite{satorras2018gnn, kim2019egnn, yang2020dpgn}). The latter methods learn task-specific metrics \cite{yoon2019tapnet, li2019ctm, qiao2019team, ye2020feat, simon2020adaptive} which can  adapt to each unseen new task. (2) Optimization-based methods \cite{ravi2017iclr, munkhdalai2017icml, finn2017icml, nichol2018reptile, rusu2019leo, Lee2019cvpr} aim to meta-learn an optimizer. Specifically, MAML \cite{finn2017icml} was proposed to learn a good model initialization with seen class data and then quickly adapt it on novel class tasks. Reptile \cite{nichol2018reptile} further simplified MAML, and MetaOptNet \cite{Lee2019cvpr} enhanced MAML by replacing the linear classifier with an SVM. (3) Hallucination-based methods \cite{hariharan2017iccv, wang2018imaginary, schwartz2018delta, zhang2019cvpr, li2020afhn} aim to learn generators from seen class samples, which are then applied during meta-testing by hallucinating new samples/features using the few shots from unseen classes. (4) Prediction-based methods \cite{qi2018imprint, qiao2018cvpr, gidaris2019wdae, guo2020attnweights} directly learn to utilize a few labeled samples to predict the parameters of neural networks for few-shot classification.

The state-of-the-art FSL results are mostly achieved by methods from the first group \cite{allen2019imp,li2019dn4,afrasiyabi2020align,ye2020feat,zhang2020deepemd}, especially those based on ProtoNet \cite{snell2017nips}. Our CPLAE is also an embedding-based method based on ProtoNet. However, armed with augmented embedding (AE) and additionally introducing a support set prototype centered loss, our model is more capable of dealing with the limited training data in FSL, resulting in superior performance (see Sec.~\ref{sec:exp}). 

\noindent\textbf{Data Augmentation for FSL.}~Several recent works \cite{hsu2019unsupervised, khodadadeh2019nips, antoniou2019assume, qin2020unsupervised, gidaris2019selfsup, su2020when, mangla2020wacv} have utilized data augmentation for meta-learning based FSL. \cite{hsu2019unsupervised, khodadadeh2019nips, antoniou2019assume, qin2020unsupervised} focus on unsupervised FSL, where augmented data samples and their original version are used to form pseudo classes to enable supervised episodic training. For supervised FSL, \cite{gidaris2019selfsup, su2020when, mangla2020wacv} take a  multi-task learning framework where augmented data are used for auxiliary self-supervised pretext tasks (e.g., predicting the rotation angle). Our CPLAE is also a supervised FSL model, but the way data augmentation is used is very different from that in \cite{gidaris2019selfsup, su2020when, mangla2020wacv}. Specifically, for each sample, we conduct three kinds of image deformations and then input the four images (together with the original one) into a feature embedding network to obtain a concatenated augmented embedding (AE) space with higher dimensionality than the original embedding space. Different orders of concatenation are further used to formulate our contrastive prototype learning (CPL) loss/objective to boost the generalization ability of the learned embedding.

\noindent\textbf{Contrastive Learning.}~Contrastive learning (CL) has recently achieved great success in self-supervised learning \cite{oord2018cpc, tian2019cmc, ting2020CoRR, he2020moco} where augmented data creates pseudo classes so that supervised learning can be applied. This has been recently extended to supervised CL \cite{Khosla2020SCL} where given an instance as anchor, all other instances (original and augmented) of the same classes are positives and the rest as negatives. Our CPL loss is essentially also a supervised CL loss. However, there are vital differences: our anchors are prototypes from the support set and critically CL is seamlessly combined with the proposed AE with different embedding concatenation orders applied to the anchor and positives/negatives respectively to challenge the generalization ability of the learned embedding. Note that ProtoTransfer \cite{medina2020self} also exploits CL for FSL, but under the unsupervised setting only, rather than our supervised FSL problem. 

\section{Methodology}

\subsection{Problem Definition}

Let $\mathcal{C}_s$ denote a set of seen classes and $\mathcal{C}_u$ a set of unseen classes, where $\mathcal{C}_s \cap \mathcal{C}_u = \emptyset$. We are given a large sample set $\mathcal{D}_s = \{ (x_i, y_i) | y_i \in \mathcal{C}_s, i = 1, \cdots, N_s\}$ from $\mathcal{C}_s$, and a few-shot sample set $\mathcal{D}_u = \{ (x_i, y_i) | y_i \in \mathcal{C}_u, i = 1, \cdots, N_u\}$ from $\mathcal{C}_u$, where $x_i$ is the $i$-th image in $\mathcal{D}_s$ (or $\mathcal{D}_u$), $y_i$ is the class label of $x_i$, and $N_s$ (or $N_u$) is the number of images in $\mathcal{D}_s$ (or $\mathcal{D}_u$). Particularly, for the $k$-shot sample set $\mathcal{D}_u$, $N_u = k |\mathcal{C}_u|$ (i.e., each class has $k$ labeled images). A test set $\mathcal{D}_t$ from $\mathcal{C}_u$ is also given, where $\mathcal{D}_u \cap \mathcal{D}_t = \emptyset$. The goal of few-shot learning (FSL) is to predict the labels of test images in $\mathcal{D}_t$ by exploiting $\mathcal{D}_s$ and $\mathcal{D}_u$ for training.

\subsection{FSL with Augmented Embeddings}

Most FSL methods \cite{finn2017icml, snell2017nips, satorras2018gnn, sung2018cvpr, Lee2019cvpr, kim2019egnn, ye2020feat} adopt episodic training on the set of seen class samples $\mathcal{D}_s$ and evaluate their models over few-shot classification tasks (i.e., episodes) sampled from the unseen classes $\mathcal{C}_u$. To form an $n$-way $k$-shot episode $e = (\mathcal{S}, \mathcal{Q})$, we first randomly sample a set of $n$ classes $\mathcal{C}$ from $\mathcal{C}_s$ (or $\mathcal{C}_u$), and then generate a support set $\mathcal{S} = \{ (x_i, y_i) | y_i \in \mathcal{C}, i = 1, \cdots, n \times k \}$ and a query set $\mathcal{Q} = \{ (x_i, y_i) | y_i \in \mathcal{C}, i = 1, \cdots, n \times q \}$ ($\mathcal{S} \cap \mathcal{Q} = \emptyset$) by sampling $k$ support and $q$ query samples from each class in $\mathcal{C}$, respectively.

We adopt Prototypical Network (ProtoNet) \cite{snell2017nips} as our baseline, which has a feature embedding network and a non-parametric nearest-neighbor classifier. ProtoNet thus only meta-learns the parameters of the embedding network. In each episode, it computes the mean feature embedding of support samples for each class $c \in \mathcal{C}$ as the prototype $\mathbf{p}_c$:
\begin{equation}
\mathbf{p}_c = \frac{1}{k} \sum_{(x_i, y_i) \in \mathcal{S}} f_\phi(x_i) \cdot I(y_i = c),
\end{equation}
where $f_\phi$ denotes the embedding network parameterized by $\phi$ with an output dimension $D$, and $I$ denotes the indicator function with its output being 1 if the input is true or 0 otherwise. Once the class prototypes are obtained from the support set, the distance of each query set sample to these prototypes are computed to construct a query centered cross-entropy loss for meta-learning $f_\phi$. 

\begin{figure*}[t]
\centering
\includegraphics[width=0.99\linewidth]{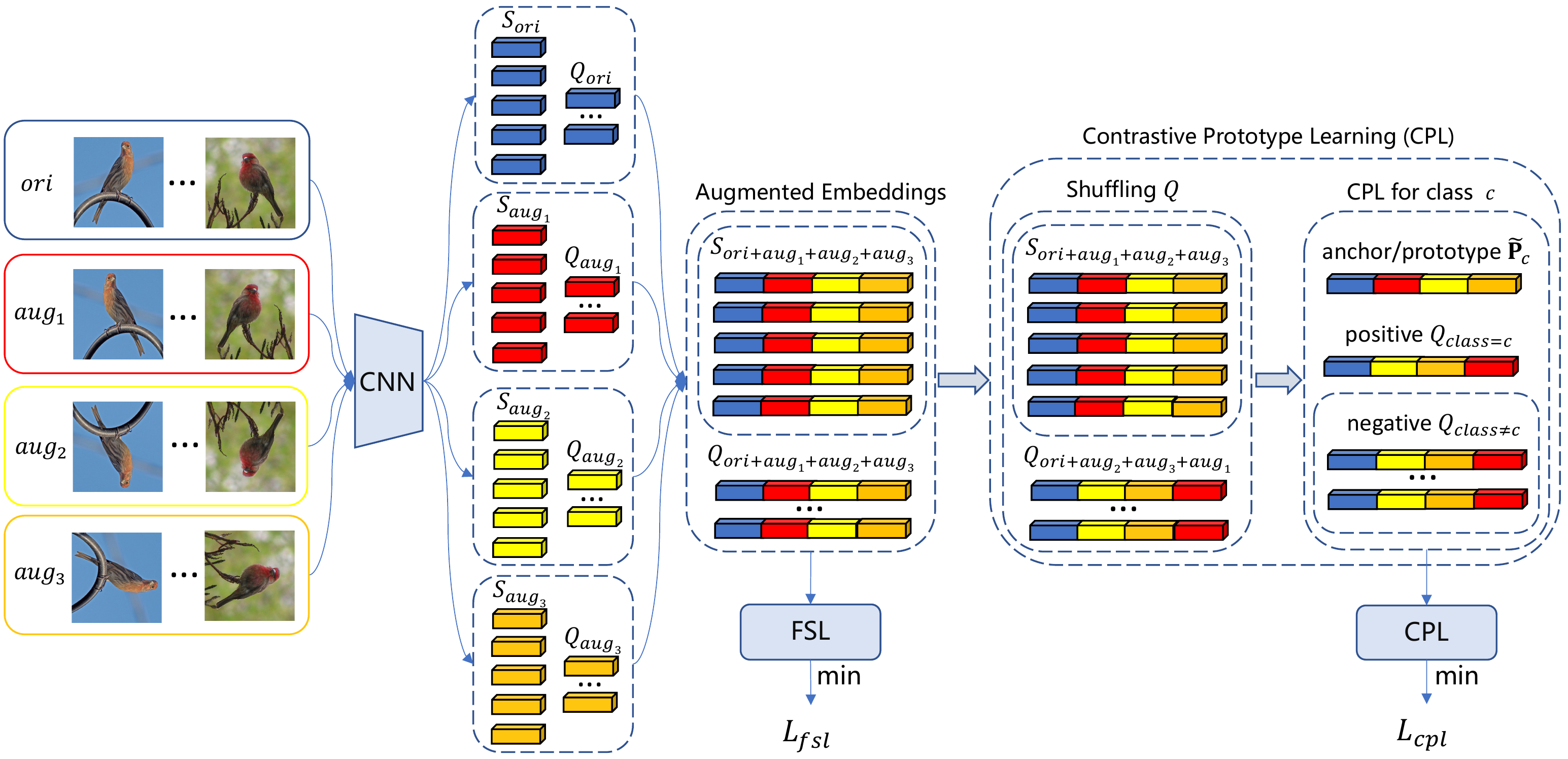}
\caption{Illustration of our proposed CPLAE model. For each original episode, we conduct three data augmentation methods to generate its three extended episodes. Concretely, samples/embeddings with the subscript $ori$ denote the original ones, while samples/embeddings with the subscripts $aug_1$, $aug_2$, and $aug_3$ are obtained by \emph{Horizontal Flip}, \emph{Vertical Flip}, and \emph{Rotation 270$^\circ$}, respectively. With the augmented embeddings by sample-wise integration, we devise two supervised losses: (1) A standard FSL loss is defined over the augmented embeddings. (2) By shuffling the concatenation order of augmented queries, a novel CPL loss is defined with prototypes as anchors.}
\label{fig:pipeline}
\vspace{-0.0in}
\end{figure*}

To deal with the lack of training data in each episode, we first apply three data augmentation $g^{(1)}$, $g^{(2)}$, and $g^{(3)}$ (e.g., horizontal flip, vertical flip, and rotations) to each image $x_i$ in $\mathcal{S} \cup \mathcal{Q}$ and obtain the corresponding feature embeddings $f_\phi^{(j)}(x_i) = f_\phi(g^{(j)}(x_i))$ ($j = 1, 2, 3$). Together with the feature embedding of the original image, our augmented embedding can be obtained by concatenating the four embeddings (see Figure~\ref{fig:pipeline}):
\begin{equation}
\tilde{f}_\phi(x_i) = A \left( f_\phi(x_i), f_\phi^{(1)}(x_i), f_\phi^{(2)}(x_i), f_\phi^{(3)}(x_i) \right),
\label{eq:aug}
\end{equation}
where $A$ is an integration function. In this work, we employ the self-attention mechanism \cite{Lin2017iclr, vaswani2017transformer} to update the four input vectors, followed by channel-wise concatenation in order, resulting in a $4D$-dimensional augmented embedding. We thus have $\tilde{f}_\phi(x_i) \in \mathbb{R}^{4D}$. In this AE space, our prototype for each class $c \in \mathcal{C}$ is obtained as:
\begin{equation}
\mathbf{\tilde{p}}_c = \frac{1}{k} \sum_{(x_i, y_i) \in \mathcal{S}} \tilde{f}_\phi(x_i) \cdot I(y_i = c).
\end{equation}

For each query sample, by computing the distances to the prototypes, we can formulate the few-shot classification loss over each episode as:
\begin{equation}
L_{fsl} = \frac{1}{q} \hspace{-3pt} \sum_{(x_i, y_i) \in \mathcal{Q}} \hspace{-3pt} -\log \frac{\exp(-d(\tilde{f}_\phi(x_i), \mathbf{\tilde{p}}_{y_i}))}{\sum_{c \in \mathcal{C}} \exp(-d(\tilde{f}_\phi(x_i), \mathbf{\tilde{p}}_c))},
\label{eq:loss_fsl}
\end{equation}
where $d(\cdot, \cdot)$ denotes the Euclidean distance between two embeddings. Note that this is a conventional FSL loss formulated from a query-centered view. The distribution of the full query set with respect to each prototype is not exploited to regularize the learned feature embedding. This can be achieved by a contrastive prototype learning (CPL) loss formulated from a support-set prototype-centered view.

\subsection{Contrastive Prototype Learning (CPL)}

Our CPL loss is a supervised contrastive learning loss. Different from the conventional un-/self-supervised CL loss, our CPL utilizes the class labels of samples in each episode to construct a few-shot supervised contrastive learning model with augmented embeddings. As illustrated in Figure~\ref{fig:pipeline}, our main idea of CPL is that: for each class $c \in \mathcal{C}$, we take the prototype $\mathbf{\tilde{p}}_c$ as the anchor, with queries from class $c$ being positive examples and queries from the other classes being negative examples.

Following the common practice in CL where positive examples contain augmented versions of the same training instance/sample, we also adopt augmentation to enrich the training data. However, there is a vital difference: we use the same set of augmentations but vary their concatenation order to produce more nuanced perturbations in the AE space.  Concretely, we shuffle the order of the three augmentations and obtain a shuffled augmented embedding $\hat{f}_{\phi}(x_i)$ of each query sample $x_i \in \mathcal{Q}$ for CPL:
\begin{equation}
\hat{f}_\phi(x_i) = A \left( f_\phi(x_i), f_\phi^{(2)}(x_i), f_\phi^{(3)}(x_i), f_\phi^{(1)}(x_i) \right).
\label{eq:aug_shuf}
\end{equation}
Note that the shuffling is only applied to the query samples, not to the support samples/anchors. Also note that in the shuffled concatenation, the original image's embedding $f_{\phi}(x_i)$ remain in the first place of $\hat{f}_\phi(x_i)$. We found empirically that once that is fixed, how exactly the other three embeddings are shuffled makes little difference (see the supplementary material for more details).

For each class $c \in \mathcal{C}$, let $\mathcal{P}^{(c)} = \{ (x_i, y_i) \in \mathcal{Q} | y_i = c, i = 1, \cdots, q \}$ denote the set of positive examples. We then compute the similarity between the anchor/prototype $\mathbf{\tilde{p}}_c$ and each $x_i$ in $\mathcal{P}^{(c)}$ as follows:
\begin{equation}
sim^{(pos)}_{c, i} = \exp( \text{cos}( \mathbf{\tilde{p}}_c, h(\hat{f}_\phi(x_i)) ) / T ),
\label{eq:sim_pos}
\end{equation}
where $h(\cdot)$ is a small neural network projection head that maps representations/emdeddings to the space where the contrastive loss is applied (as in \cite{ting2020CoRR}), $\text{cos}(\cdot, \cdot)$ computes the cosine similarity between two vectors, and $T$ is the temperature parameter. For each positive example $(x_i, y_i) \in \mathcal{P}^{(c)}$, we first randomly sample $m$ ($m \le q$) query samples from each of the other classes to form the set of negative examples $\mathcal{N}^{(c)}_i = \{ (x_t, y_t) \in \mathcal{Q} | y_t \ne c, t = 1, \cdots, m(n-1) \}$. We then obtain the similarities for all negative examples:
\begin{equation}
sim^{(neg)}_{c, i} = \sum_{(x_t, y_t) \in \mathcal{N}^{(c)}_i} \exp( \text{cos} (\mathbf{\tilde{p}}_c, h(\hat{f}_\phi(x_t)) ) / T).
\label{eq:sim_neg}
\end{equation}
The contrastive loss used for CPL is finally given by:
\begin{equation}
L_{cpl} = \frac{1}{nq} \sum_{c \in \mathcal{C}} \sum_{(x_i, y_i) \in \mathcal{P}^{(c)}} \hspace{-6pt}- \log \frac{sim^{(pos)}_{c, i}}{sim^{(pos)}_{c,i} + sim^{(neg)}_{c,i}}.
\label{eq:loss_cpl}
\end{equation}

From this formulation, it is clear that compared to the popular unsupervised CL \cite{ting2020CoRR}, our CPL loss is supervised in that it utilizes the class labels of samples. Compared with existing supervised contrastive losses such as triplet loss \cite{Florian2015cvpr}, its improved version N-pair loss \cite{Kihyuk2016nips}, and the more recent supervised CL loss \cite{Khosla2020SCL}, our CPL loss has two main differences: (1) Our CPL is designed for FSL which takes class prototypes as anchors, while \cite{Florian2015cvpr, Kihyuk2016nips, Khosla2020SCL} take samples as anchors. (2) The contrastive learning is conducted in an AE space with perturbations on the concatenation orders of the augmented feature embeddings to boost the generalization ability of the learned feature embedding.

\subsection{Learning Objectives for CPLAE}

\renewcommand{\algorithmicrequire}{\textbf{Input:}}
\renewcommand{\algorithmicensure}{\textbf{Output:}}
\begin{algorithm}[t]
\caption{CPLAE for FSL}
\label{alg:CPL}
\begin{algorithmic}[1]
\REQUIRE Our CPLAE model $M_{\Theta}$ ($\Theta$ is the set of parameters) \\
\quad~~The seen class training set $\mathcal{D}_s$ \\
\quad~~The hyper-parameters $\lambda$, $T$, $m$
\ENSURE The learned $M_{\Theta}^*$
\FORALL{iteration = 1, 2, $\cdots$, MaxIteration}
\STATE Sample an $n$-way $k$-shot episode $e$ from $\mathcal{D}_s$;
\STATE Obtain $\tilde{f}_\phi(x)$ as the augmented embedding for each sample $x$ from $e$ with Eq.~(\ref{eq:aug});
\STATE Compute $L_{fsl}$ with Eq.~(\ref{eq:loss_fsl});
\STATE Obtain $\hat{f}_\phi(x)$ as the shuffled augmented embedding for each query sample $x$ from $\mathcal{Q}$ with Eq.~(\ref{eq:aug_shuf});
\STATE Compute $L_{cpl}$ with Eq.~(\ref{eq:loss_cpl});
\STATE Compute the total loss $L_{total}$ with Eq.~(\ref{eq:loss_total});
\STATE Compute the gradients $\nabla_{M_{\Theta}} L_{total}$;
\STATE Update $M_{\Theta}$ using stochastic gradient descent;
\ENDFOR
\STATE \textbf{return} the found best $M_{\Theta}^*$.
\end{algorithmic}
\end{algorithm}

In each training iteration, we randomly sample one $n$-way $k$-shot $q$-query episode $e = (\mathcal{S}, \mathcal{Q})$. For each instance/sample in $e$, we apply three different data augmentation methods on it, and then integrate the obtained four feature embeddings into one augmented embedding. The few-shot classification loss $L_{fsl}$ is computed with the augmented embeddings according to Eq.~(\ref{eq:loss_fsl}). Moreover, for all query samples, we shuffle the integrating order of their augmented embeddings and compute the CPL loss $L_{cpl}$ in Eq.~(\ref{eq:loss_cpl}). The total learning objective for the proposed Contrastive Prototype Learning with Augmented Embeddings (CPLAE) model is finally stated as:
\begin{equation}
L_{total} = L_{fsl} + \lambda L_{cpl},
\label{eq:loss_total}
\end{equation}
where $\lambda$ is used to balance the importance of the FSL and CPL losses. In this work, $\lambda$ is empirically set to 0.1. Our full CPLAE algorithm is outlined in Algorithm~\ref{alg:CPL}. Once learned, with the optimal model found by our CPLAE algorithm, we randomly sample multiple $n$-way $k$-shot meta-test episodes from $\mathcal{C}_u$ for performance evaluation.

\begin{table*}[t]
\centering
\scalebox{0.9}{
\tabcolsep12pt
\begin{tabular}{lccccc}
\Xhline{1pt}
& & \multicolumn{2}{c}{\textbf{\emph{mini}ImageNet}} & \multicolumn{2}{c}{\textbf{\emph{tiered}ImageNet}} \\
\textbf{Method} & \textbf{Backbone} & \textbf{5-way 1-shot} & \textbf{5-way 5-shot} & \textbf{5-way 1-shot} & \textbf{5-way 5-shot}\\
\hline \hline
MatchingNet~(NeurIPS'16) \cite{oriol2016nips}  & Conv4-64 & $43.56\pm0.84$ & $55.31\pm0.73$ & -- & -- \\
ProtoNet$^{\dagger}$~(NeurIPS'17) \cite{snell2017nips} & Conv4-64 & $52.79\pm0.45$ & $71.23\pm0.36$ & $53.82\pm0.48$ & $71.77\pm0.41$ \\
MAML~(ICML'17) \cite{finn2017icml} & Conv4-64 & $48.70\pm1.84$ & $63.10\pm0.92$ & $51.67\pm1.81$ & $70.30\pm0.08$ \\
RelationNet~(CVPR'18) \cite{sung2018cvpr} & Conv4-64 & $50.40\pm0.80$ & $65.30\pm0.70$ & $54.48\pm0.93$ & $71.32\pm0.78$ \\
IMP~(ICML'19) \cite{allen2019imp}  & Conv4-64 & $49.60\pm0.80$ & $68.10\pm0.80$ & - & - \\
DN4~(CVPR'19) \cite{li2019dn4} & Conv4-64 & $51.24\pm0.74$ & $71.02\pm0.64$ & -- & -- \\DN
PARN~(ICCV'19) \cite{wu2019parn} & Conv4-64 & $55.22\pm0.84$ & $71.55\pm0.66$ & -- & -- \\
PN+rot~(ICCV'19) \cite{gidaris2019selfsup} & Conv4-64 & $53.63\pm0.43$ & $71.70\pm0.36$ & -- & -- \\
CC+rot~(ICCV'19) \cite{gidaris2019selfsup} & Conv4-64 & $54.83\pm0.43$ & $71.86\pm0.33$ & -- & -- \\
Centroid~(ECCV'20) \cite{afrasiyabi2020align} & Conv4-64 & $53.14\pm1.06$ & $71.45\pm0.72$ & -- & -- \\
Neg-Cosine~(ECCV'20) \cite{liu2020negative} & Conv4-64 & $52.84\pm0.76$ & $70.41\pm0.66$ & -- & -- \\
FEAT~(CVPR'20) \cite{ye2020feat} & Conv4-64 & $55.15 \pm 0.20$ & $71.61 \pm 0.16$ & -- & -- \\
\hline
CPLAE (ours) & Conv4-64 & $\bf56.83\pm0.44$ & $\bf74.31\pm0.34$ & $\bf58.23\pm0.49$ & $\bf75.12\pm0.40$ \\
\hline
\hline
ProtoNet$^{\dagger}$~(NeurIPS'17) \cite{snell2017nips} & Conv4-512 & $53.52\pm0.43$ & $73.34\pm0.36$ & $55.52\pm0.48$ & $74.07\pm0.40$ \\
MAML~(ICML'17) \cite{finn2017icml} & Conv4-512 & $49.33\pm0.60$ & $65.17\pm0.49$ & $52.84\pm0.56$ & $70.91\pm0.46$ \\
Relation Net~(CVPR'18) \cite{sung2018cvpr} & Conv4-512 & $50.86\pm0.57$ & $67.32\pm0.44$ & $54.69\pm0.59$ & $72.71\pm0.43$ \\
PN+rot~(ICCV'19) \cite{gidaris2019selfsup} & Conv4-512 & $56.02\pm0.46$ & $74.00\pm0.35$ & -- & -- \\
CC+rot~(ICCV'19) \cite{gidaris2019selfsup} & Conv4-512 & $56.27\pm0.43$ & $74.30\pm0.33$ & -- & -- \\
\hline
CPLAE (ours) & Conv4-512 & $\bf57.46\pm0.43$ & $\bf75.69\pm0.33$ & $\bf61.56\pm0.50$ & $\bf80.03\pm0.38$ \\
\hline
\hline
ProtoNet$^{\dagger}$~(NeurIPS'17) \cite{snell2017nips} & ResNet-12 & $62.41\pm0.44$ & $80.49\pm0.29$ & $69.63\pm0.53$ & $84.82\pm0.36$ \\
TADAM~(NeurIPS'18) \cite{oreshkin2018tadam} & ResNet-12 & $58.50\pm0.30$ & $76.70\pm0.38$  & --  & --  \\
MetaOptNet~(CVPR'19) \cite{Lee2019cvpr} & ResNet-12 & $62.64\pm0.61$ & $78.63\pm0.46$ & $65.99\pm0.72$  &  $81.56\pm0.63$\\
MTL~(CVPR'19) \cite{sun2019meta} & ResNet-12 & $61.20\pm1.80$ & $75.50\pm0.80$ & $65.62 \pm1.80$ & $80.61\pm0.90$ \\
AM3~(NeurIPS'19) \cite{xing2019am3} & ResNet-12 & $65.21\pm0.49$ & $75.20\pm0.36$ & $67.23\pm0.34$ & $78.95\pm0.22$ \\
Shot-Free~(ICCV'19) \cite{ravichandran2019few} & ResNet-12 & $59.04\pm0.43$ & $77.64\pm0.39$ & $66.87\pm0.43$ & $82.64\pm0.43$\\
Neg-Cosine~(ECCV'20) \cite{liu2020negative} & ResNet-12 & $63.85\pm0.81$ & $81.57\pm0.56$ & -- & -- \\
Distill~(ECCV'20) \cite{tian2020rethink} & ResNet-12 & $64.82\pm0.60$ & $82.14\pm0.43$ & $71.52\pm0.69$ & $86.03\pm0.49$ \\
DSN-MR~(CVPR'20) \cite{simon2020adaptive} & ResNet-12 & $64.60\pm0.72$ & $79.51\pm0.50$ & $67.39\pm0.82$ & $82.85 \pm0.56$ \\
DeepEMD~(CVPR'20) \cite{zhang2020deepemd} & ResNet-12 & $65.91\pm0.82$ & $82.41\pm0.56$ & $71.16\pm0.87$ & $86.03\pm0.58$ \\
FEAT~(CVPR'20) \cite{ye2020feat} & ResNet-12 & $66.78\pm0.20$ & $82.05\pm0.14$ & $70.80\pm0.23$ & $84.79\pm0.16$ \\
\hline
CPLAE (ours) & ResNet-12 & $\bf67.46\pm0.44$ & $\bf83.22\pm0.29$ & $\bf72.23\pm0.50$ & {$\bf87.35\pm0.34$} \\
\Xhline{1pt}
\end{tabular}
}
\caption{Comparative results of standard FSL on the two benchmark datasets. The average 5-way few-shot classification accuracies (\%, top-1) along with the 95\% confidence intervals are reported.}
\label{tab:main}
\vspace{-0.1in}
\end{table*}

\section{Experiments}
\label{sec:exp}
\subsection{Datasets and Settings}

\noindent\textbf{Datasets.} We select three widely-used benchmarks for evaluation: \emph{mini}ImageNet \cite{oriol2016nips}, \emph{tiered}ImageNet \cite{ren2018fs-ssl}, and CUB-200-2011 \cite{CUB-200-2011}. The \emph{mini}ImageNet dataset contains 100 classes from  ILSVRC-12 \cite{Russakovsky2015ImageNet}, with each class having 600 images. We split it into 64 training classes, 16 validation classes, and 20 test classes, as in \cite{ravi2017iclr}. The \emph{tiered}ImageNet dataset is a larger subset of ILSVRC-12, which consists of 608 classes and 779,165 images in total. We split it into 351 training classes, 97 validation classes, and 160 test classes, as in \cite{ren2018fs-ssl}. Different from the aforementioned two, CUB-200-2011 is a fine-grained classification dataset consisting of 11,778 images from 200 different bird classes. The 200 classes are divided into 100, 50, 50 classes for training, validation and testing, respectively. All images of the  datasets are resized to $84 \times 84$ before being inputted into the feature embedding networks (i.e., CNNs).

\noindent\textbf{Evaluation Protocols.} We make evaluation under 5-way 5-shot/1-shot as in previous works. Each episode has 5 randomly sampled classes from the test split, each of which contains 5 shots (or 1 shot) and 15 queries. We thus have $n = 5$, $k = 5~\mathrm{or}~1$, and $q = 15$. Note that since data augmentations can be performed easily (in a fully unsupervised way), we also adopt the augmented embeddings for all images during evaluation, and keep the integration order as in Eq.~(\ref{eq:aug}) (i.e., no shuffling is involved). We report average 5-way classification accuracy (\%, top-1) over 2,000 meta-test episodes along with the 95\% confidence interval.

\noindent\textbf{Feature Embedding Networks.} We adopt three backbones as the feature extractors $f_\phi$: Conv4-64 \cite{oriol2016nips}, Conv4-512, and ResNet-12 \cite{he2016resnet}. They all take the same input image size of $84 \times 84$. Particularly, both Conv4-64 and Conv4-512 consist of 4 convolutional layers: the first three layers are exactly the same but the last layer has different numbers of out channels in the two backbones. Since we use an average pooling layer after the last convolutional layer for each backbone, the output feature dimensions of Conv4-64, Conv4-512, and ResNet-12 are 64, 512, and 640, respectively. We pre-train all three backbones on the training split of each dataset to accelerate the training process, as in \cite{zhang2020deepemd, ye2020feat, simon2020adaptive}. With the pre-trained backbones, our CPLAE is then applied in the meta-training stage. For ResNet-12, the stochastic gradient descent (SGD) optimizer is employed with the initial learning rate of 1e-4, the weight decay of 5e-4, and the Nesterov momentum of 0.9. For Conv4-64 and Conv4-512, the Adam optimizer \cite{kingma2015adam} is adopted (instead of SGD) with the initial learning rate of 1e-4.

\noindent\textbf{Implementation Details.}  In all experiments, our CPLAE is trained for 100 epochs with 100 episodes per epoch, and the learning rate is halved every 20 epochs. The hyper-parameters are selected according to the validation performance of our algorithm. Particularly, for each class $c \in \mathcal{C}$, we sample 6 negative examples (i.e., $m = 6$) from each class in $\mathcal{C}\setminus\{c\}$ for every positive example. While computing the similarity between the anchor and each positive example in Eq.~(\ref{eq:sim_pos}), the temperature $T$ is set to 1. The code and models will be released soon.

\subsection{Main Results}

For comprehensive comparison, we select a variety of latest/state-of-the-art FSL methods \cite{ye2020feat,zhang2020deepemd,simon2020adaptive,tian2020rethink,liu2020negative,afrasiyabi2020align} as well as the strongest SSL+FSL method CC+rot~\cite{gidaris2019selfsup} as the competitors, in addition to the classic/representative baselines (e.g., ProtoNet and MAML). The comparative results of standard/conventional FSL on \emph{mini}ImageNet \cite{oriol2016nips} and \emph{tiered}ImageNet are provided in Table~\ref{tab:main}. More results on fine-grained CUB and cross-domain \emph{mini}ImageNet$\rightarrow$CUB are presented in the supplementary material due to space constraint. Note that we re-implement our main baseline (i.e., ProtoNet \cite{snell2017nips}, denoted with $^\dag$) with the same hyper-parameter during training for fair comparison.

We can observe that: (1) With the same backbone (out of the three ones), our CPLAE achieves new state-of-the-art on all datasets under both 1-shot and 5-shot settings, validating the effectiveness of CPL with augmented embeddings (AE). This suggests that our CPLAE has the strongest generalization ability thanks to the introduction of AE and the use of prototype centered contrastive learning. (2) Impressively, our CPLAE with Conv4-64 even outperforms the state-of-the-art competitors with Conv4-512 in all cases. Since the performance achieved by  our CPLAE even surpasses that of the strongest competitor CC+rot~\cite{gidaris2019selfsup} that also utilised data augmentation but with a stronger backbone, our results validate our novel way of using augmentation (self-attention + concatenation) and highlight the importance of support-centered meta-learning loss. (3) The improvements obtained by our CPLAE over the baseline ProtoNet$^\dag$ range from 2.3\% to 6.0\%, providing direct evidence that both the proposed augmented embeddings and contrastive prototype learning bring significant benefits to FSL (further evidence is provided in ablation study results shortly). 

\subsection{Further Evaluation}
\label{sub:fe}

\noindent\textbf{Ablation Study.} Our full CPLAE model is trained with two losses: the FSL loss $L_{fsl}$ and the CPL loss $L_{cpl}$ (see Eq.~(\ref{eq:loss_total})). For $L_{fsl}$, we adopt an augmented embedding for each sample, which is obtained by integrating four feature vectors (one from the original image and three from its augmented ones). For $L_{cpl}$, we devise a novel supervised contrastive loss with the shuffling operation of query samples. To demonstrate the contribution of each main component, we conduct ablative experiments on \emph{mini}ImageNet in Table~\ref{tab:ablation}, where Conv4-64 is adopted as the backbone. Four methods are compared: (1) ProtoNet$^\dag$: our re-implementation of ProtoNet \cite{snell2017nips}. (2) ProtoNet$^\dag$+AE: ProtoNet trained with augmented embeddings (i.e., trained with only $L_{fsl}$ in Eq.~(\ref{eq:loss_total})). (3) CPLAE (no-shuffling): Our CPLAE model trained with the total loss in Eq.~(\ref{eq:loss_total}) but without the shuffling operation. (4) CPLAE: our full CPLAE model. The ablation study results in Table~\ref{tab:ablation} show that the augmented embeddings lead to 2--3\% improvements (see ProtoNet$^\dag$+AE vs. ProtoNet$^\dag$), and our proposed CPL further improves the performance by about 1\% (see CPLAE vs. ProtoNet$^\dag$+AE). In addition, the comparison CPLAE vs. CPLAE (no shuffling) demonstrates the importance of the shuffling operation for our CPL.

\begin{table}[t]
\centering
\scalebox{0.95}{
\begin{tabular}{lcc}
\Xhline{1pt}
\textbf{Method} & \textbf{5-way 1-shot} & \textbf{5-way 5-shot} \\
\hline \hline
ProtoNet$^\dag$      & $52.79\pm0.45$ & $71.23\pm0.36$ \\
ProtoNet$^\dag$+AE   & $55.89\pm0.43$ & $73.43\pm0.35$ \\
CPLAE~(no shuffling) & $56.04\pm0.44$ & $73.75\pm0.35$ \\
CPLAE                & $\bf56.83\pm0.44$ & $\bf74.31\pm0.34$ \\
\Xhline{1pt}
\end{tabular}}
\caption{Ablation study results for our full CPLAE model on \emph{mini}ImageNet (with Conv4-64 being the backbone). }\label{tab:ablation}
\end{table}

\begin{table}[t]
\centering
\scalebox{0.85}{
\tabcolsep2pt
\begin{tabular}{cccccc}
\Xhline{1pt}
\multicolumn{5}{c}{\textbf{Data Augmentation Methods}} & \\
\textbf{Horizontal} & \textbf{Vertical} & \textbf{Rotation} & \textbf{Rotation} & \textbf{Rotation} & \multirow{2}{*}{\textbf{5-way 5-shot}} \\
\textbf{Flip} & \textbf{Flip} & \textbf{90$^\circ$} & \textbf{180$^\circ$} & \textbf{270$^\circ$} \\
\hline\hline
$\times$ & $\times$ & $\times$ & $\times$ & $\times$ & $71.23\pm0.36$ \\
$\checkmark$ & $\checkmark$ & & & & $72.96\pm0.35$ \\
$\checkmark$ & $\checkmark$ & $\checkmark$ & & & $73.41\pm0.35$ \\
$\checkmark$ & $\checkmark$ & & $\checkmark$ & & $73.34\pm0.35$ \\
$\checkmark$ & $\checkmark$ & & & $\checkmark$ & $\mathbf{73.43\pm0.35}$ \\
$\checkmark$ & & $\checkmark$ & $\checkmark$ & & $73.40\pm0.34$ \\
$\checkmark$ & & $\checkmark$ & & $\checkmark$ & $73.34\pm0.34$ \\
$\checkmark$ & & & $\checkmark$ & $\checkmark$ & $73.28\pm0.35$ \\
& $\checkmark$ & $\checkmark$ & $\checkmark$ & & $72.62\pm0.35$ \\
& $\checkmark$ & $\checkmark$ & & $\checkmark$ & $73.14\pm0.35$ \\
& $\checkmark$ & & $\checkmark$ & $\checkmark$ & $72.90\pm0.35$ \\
& & $\checkmark$ & $\checkmark$ & $\checkmark$ & $73.14\pm0.35$ \\
$\checkmark$ & $\checkmark$ & $\checkmark$ & & $\checkmark$ & $72.77\pm0.36$ \\
$\checkmark$ & $\checkmark$ & $\checkmark$ & $\checkmark$ & & $73.10\pm0.35$ \\
\Xhline{1pt}
\end{tabular}}
\caption{Comparison among different choices of the data augmentation methods under the 5-way 5-shot setting on \emph{mini}ImageNet. Conv4-64 is used as the feature extractor.}
\label{tab:alter_aug}
\vspace{-0.1in}
\end{table}

\noindent\textbf{Alternative Augmentation Strategies.} In Table~\ref{tab:alter_aug}, we compare different choices of the data augmentation methods for our augmented embeddings. We select five common image deformation methods and use their combinations as the alternative data augmentation strategies. Note that the first row of Table~\ref{tab:alter_aug} involves no data augmentation (i.e., ProtoNet$^\dag$), and the rest results are obtained by ProtoNet$^\dag$+AE. Particularly, the second row means that we only use two deformation methods (i.e., the dimension of augmented embeddings in this case is $3D$), and the last two rows use four methods (i.e., the integrated embeddings are of $5D$). The comparative results in Table~\ref{tab:alter_aug} show that the combination of Horizontal Flip, Vertical Flip, and Rotation 270$^\circ$ is the best, and FSL using three deformation methods outperforms FSL using two or four. 

\begin{table}[t]
\centering
\begin{tabular}{lcc}
\Xhline{1pt}
\textbf{Method} & \textbf{5-way 1-shot} & \textbf{5-way 5-shot} \\
\hline \hline
ProtoNet$^\dag$+AE     & $55.89\pm0.43$ & $73.43\pm0.35$ \\
ProtoNet$^\dag$+AE+PT  & $55.62\pm0.45$ & $73.24\pm0.35$ \\
ProtoNet$^\dag$+AE+CL  & $55.61\pm0.44$ & $73.52\pm0.53$ \\
CPLAE                  & $\bf56.83\pm0.44$ & $\bf74.31\pm0.34$ \\
\Xhline{1pt}
\end{tabular}
\caption{Comparison to different SSL losses on \emph{mini}ImageNet (with Conv4-64 being the backbone). PT -- SSL based on pretext tasks; CL -- SSL via contrastive learning.}
\label{tab:ssl}
\end{table}

\begin{table}[t]
\centering
\scalebox{0.87}{
\begin{tabular}{lcc}
\Xhline{1pt}
\textbf{Method} & \textbf{5-way 1-shot} & \textbf{5-way 5-shot} \\
\hline \hline
ProtoNet$^\dag$+AE          & $55.89\pm0.43$ & $73.43\pm0.35$ \\
CPLAE~(w/o Proto, w/o Proj) & $56.18\pm0.44$ & $73.15\pm0.35$ \\
CPLAE~(w/o Proto, w/ Proj)  & $56.24\pm0.43$ & $73.35\pm0.35$ \\
CPLAE~(w/ Proto, w/o Proj)  & $56.31\pm0.44$ & $73.62\pm0.34$ \\
CPLAE~(w/ Proto, w/ Proj)   & $\bf56.83\pm0.44$ & $\bf74.31\pm0.34$ \\
\Xhline{1pt}
\end{tabular}}
\caption{Comparison to CPL alternatives on \emph{mini}ImageNet. Conv4-64 is used as the feature extractor.}
\label{tab:alter_cpl}
\end{table}

\begin{figure*}[t]
\centering
\includegraphics[width=0.98\linewidth]{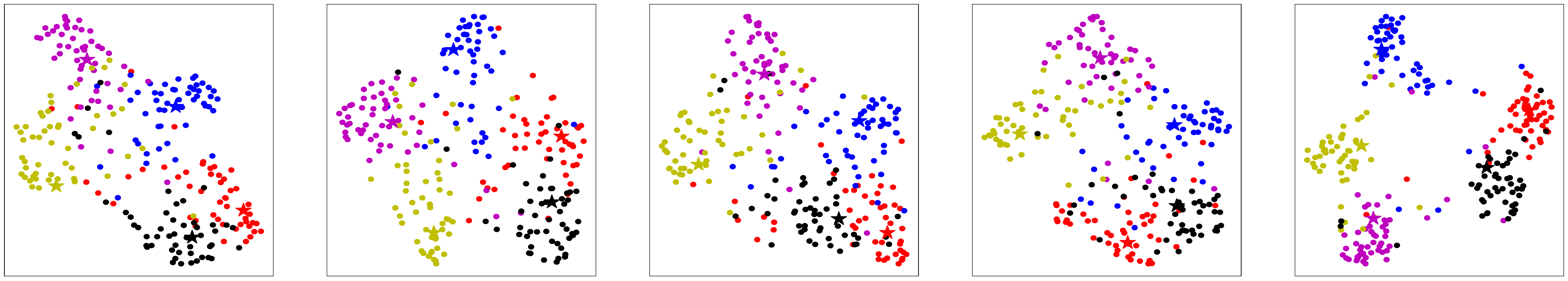}
\caption{Visualizations of data distributions of five episodes on \emph{mini}ImageNet using the UMAP algorithm \cite{UMAP}. The first four columns present the results of one original meta-test episode and its three extended ones, respectively. The last column presents the visualization using integrated embeddings. The 5-way 5-shot setting is considered, with Conv4-64 as the feature extractor. }
\label{fig:vis1}
\end{figure*}

\begin{figure*}[t]
\centering
\includegraphics[width=0.98\linewidth]{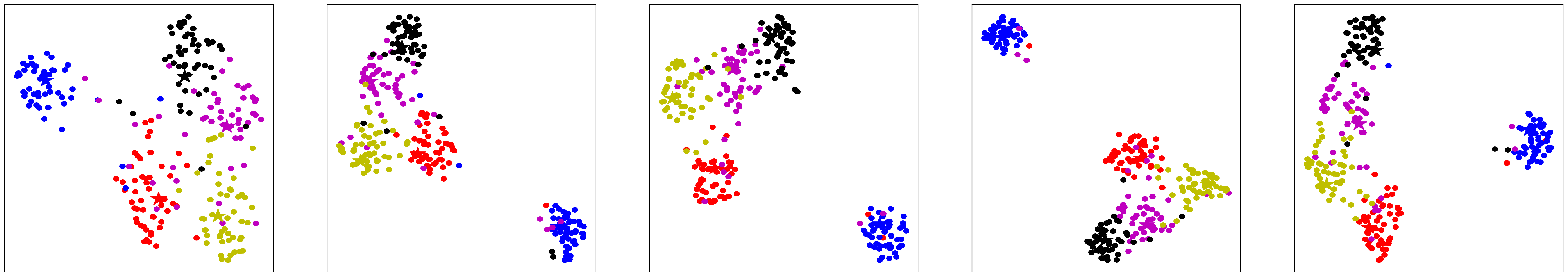}
\caption{Visualizations of data distributions of the same meta-test episode from \emph{mini}ImageNet using the UMAP algorithm \cite{UMAP} for five FSL models (from left to right): ProtoNet$^\dag$, ProtoNet$^\dag$+AE, CPLAE~(w/o Proto, w/o Proj, no shuffling), CPLAE~(no shuffling), and our CPLAE. The 5-way 5-shot setting is considered, with Conv4-64 as the feature extractor. }
\label{fig:vis2}
\vspace{-0.05in}
\end{figure*}

\noindent\textbf{Alternative SSL Losses.} Our supervised contrastive learning loss $L_{cpl}$ is inspired by the previous self-supervised learning (SSL) and contrastive learning (CL) works. To verify the effectiveness of our CPLAE model with the CPL loss, we compare it to two alternative models with SSL losses. (1) ProtoNet$^\dag$+AE+PT: an SSL loss based on the pretext task (PT) is added into ProtoNet$^\dag$+AE by predicting the augmented embeddings are shuffled or not. (2) ProtoNet$^\dag$+AE+CL: the conventional unsupervised contrastive loss is applied to ProtoNet$^\dag$+AE. Specifically, for each query sample, since we have two augmented embeddings by integration with different orders, we take each one as the anchor in turn. Naturally, the other one is treated as the positive example, while all of the rest are negative examples. The comparative results in Table~\ref{tab:ssl} demonstrate that our proposed CPL is the best choice for FSL.

\noindent\textbf{Alternative Contrastive Learning Losses.} The main differences between our proposed contrastive prototype learning (CPL) and the conventional supervised triplet loss \cite{Florian2015cvpr} (or its improved version N-pair loss \cite{Kihyuk2016nips}) are in two aspects: (1) Our CPL chooses class prototypes as anchors, but the triplet/N-pair loss takes each sample to be the anchor in turn. (2) Our CPL adopts a projection head which maps the embeddings into a latent space, but such projection is not considered in the triplet/N-pair loss. Therefore, we conduct a group of experiments to find out the contribution of the prototype-based anchors and projection head in Table~\ref{tab:alter_cpl}. Methods with `Proto' use prototypes as anchors and those with `Proj' use the projection head when computing $L_{cpl}$. We can observe that our novel integration of contrastive learning into FSL, i.e., CPLAE~(w/ Proto, w/ Proj), achieves the best results.

\subsection{Visualization Results}

When training our CPLAE model, for each sampled original episode, we form three extended episodes using three data augmentation methods (i.e., Horizontal Flip, Vertical Flip, Rotation 270$^\circ$). Therefore, the augmented embedding of each image is obtained by integrating four feature vectors from the original image and its three deformed ones. Since the three data augmentation methods are performed in a fully unsupervised way, we can also form three extended episodes for each meta-test episode. To obtain visualization results, we choose one meta-test episode and visualize it in Figure~\ref{fig:vis1}. Concretely, the first four columns present the data distributions of the original meta-test episode and its three extended episodes, while the last column presents the visualization results using the integrated embeddings. Note that all five embeddings are obtained using the same trained CPLAE model. The visualization results in Figure~\ref{fig:vis1} show that: (1) The data distributions of the first four episodes are similar, which means that our CPLAE model cannot be well learned when the augmented images are used separately. (2) The augmented embedding has the best data clustering structure, indicating that the implicit alignment of the original episode and its three extended ones is crucial for FSL with only few shots.

Based on augmented embeddings, Our CPLAE model takes the class prototypes as anchors and the shuffled embeddings of query samples as positive/negative examples. In Figure~\ref{fig:vis2}, we visualize the data distributions of the same meta-test episode obtained by five FSL models: ProtoNet$^\dag$, ProtoNet$^\dag$+AE, CPLAE~(w/o Proto, w/o Proj, no shuffling), CPLAE~(no shuffling), and our CPLAE. We can observe that: (1) CPLAE~(no shuffling) leads to better data clustering structure than CPLAE~(w/o Proto, w/o Proj, no shuffling), which suggests that the prototype-based anchors and the projection head are effective. (2) CPLAE with shuffling is better than CPLAE~(no shuffling), which indicates the necessity of using the shuffling operation for CPL.

\section{Conclusion}

We have proposed a novel contrastive prototype learning with augmented embedding (CPLAE) model to address the lack of training data problem in FSL. Different from existing embedding-based meta-learning methods, we introduce both data augmentation to form an augmented embedding space and a support set prototype centered loss to complement the conventional query centered loss. Extensive experiments on three widely used benchmarks demonstrate that our CPLAE achieves new state-of-the-art. This work shows for the first time that contrastive learning is effective under the supervised and few-shot learning setting.

{\small
\bibliographystyle{ieee_fullname}
\bibliography{CPLAE}
}

\end{document}